\documentclass[10pt,twocolumn,letterpaper]{article}

\usepackage{iccv}
\usepackage{times}
\usepackage{epsfig}
\usepackage{graphicx}
\usepackage{amsmath}
\usepackage{amssymb}
\usepackage{multirow}
\usepackage[dvipsnames]{xcolor}
\usepackage{authblk}
\usepackage[toc,page]{appendix}

\newcommand{\et}{\textit{et al. }}
\newcommand{\x}{{\bf x}}

\usepackage[pagebackref=true,breaklinks=true,letterpaper=true,colorlinks,bookmarks=false]{hyperref}

 \iccvfinalcopy 

\def\iccvPaperID{3272} 

\ificcvfinal\fi

\begin{document}

\title{Deep Constrained Dominant Sets for Person Re-identification}

\author[1]{Leulseged Tesfaye Alemu}
\author[1, 2]{Marcello Pelillo}
\author[3]{Mubarak Shah}

\affil[1]{Ca' Foscari University of Venice}
\affil[2]{ECLT, European Centre for Living Technology}
\affil[3]{CRCV, University of Central Florida}
\affil[ ]{\tt\small {leulseged.alemu@unive.it, shah@crcv.ucf.edu, pelillo@unive.it}}

\maketitle
\begin{abstract}
	In this work, we propose an end-to-end constrained clustering scheme to tackle the person re-identification (re-id) problem. Deep neural networks (DNN) have recently proven to be effective on person re-identification task. In particular, rather than leveraging solely a probe-gallery similarity, diffusing the similarities among the gallery images in an end-to-end manner has proven to be effective in yielding a robust probe-gallery affinity. However, existing methods do not apply probe image as a constraint, and are prone to noise propagation during the similarity diffusion process. To overcome this, we propose an intriguing scheme which treats person-image retrieval problem as a {\em constrained clustering optimization} problem, called deep constrained dominant sets (DCDS). Given a probe and gallery images, we re-formulate person re-id problem as finding a constrained cluster, where the probe image is taken as a constraint (seed) and each cluster corresponds to a set of images corresponding to the same person. By optimizing the constrained clustering in an end-to-end manner, we naturally leverage the contextual knowledge of a set of images corresponding to the given person-images. We further enhance the performance by integrating an auxiliary net alongside DCDS, which employs a multi-scale Resnet. To validate the effectiveness of our method we present experiments on several benchmark datasets and show that the proposed method can outperform state-of-the-art methods.
\end{abstract}

\vspace{-0.8cm}
\section{Introduction}
\begin{figure}[t]

	\begin{center}
		
		\includegraphics[width=1\linewidth ,trim=0cm 0cm 0cm 0cm,clip]{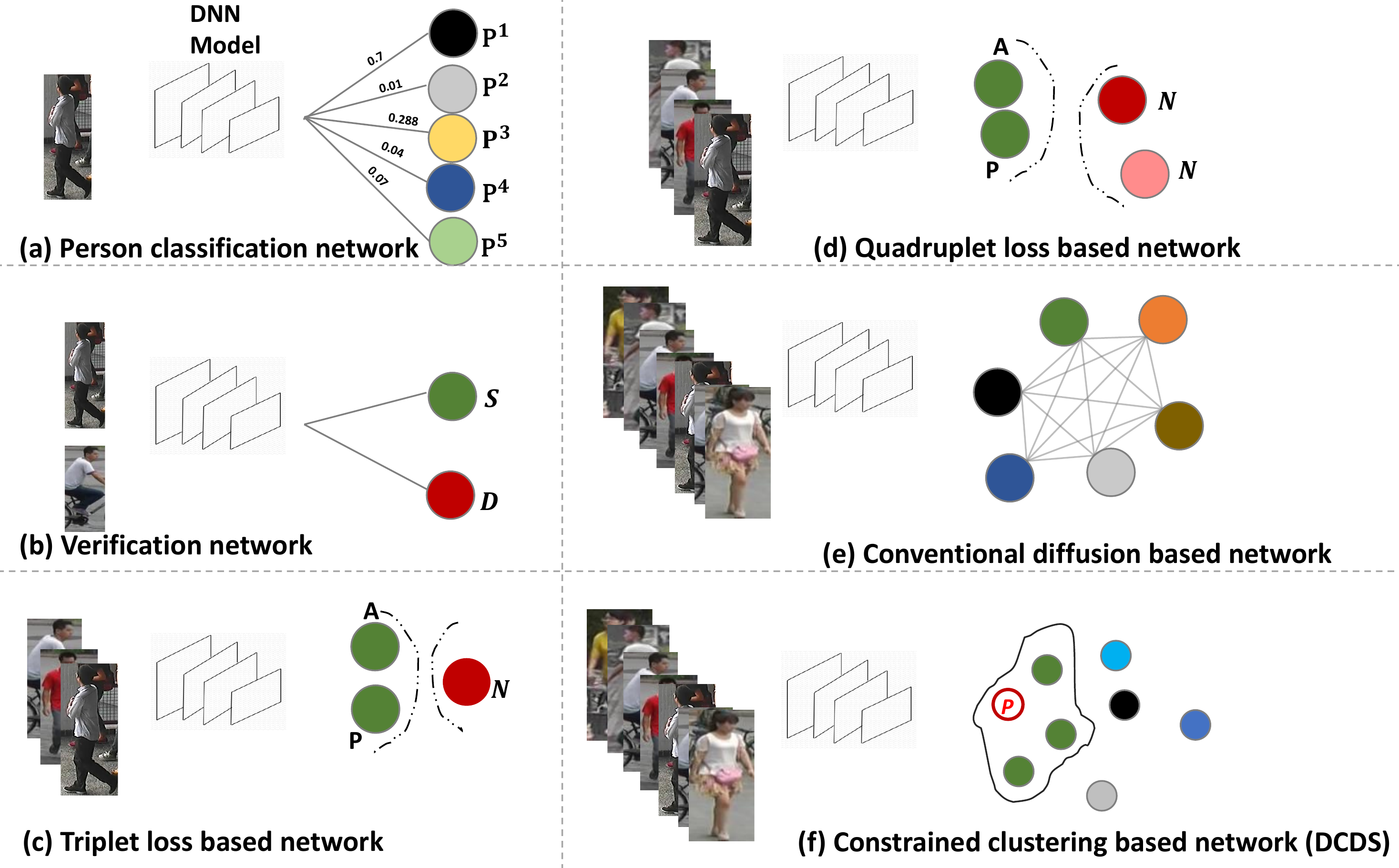}
	\end{center}
	
	\caption{Shows a variety of existing classification and similarity-based deep person re-id models. (a) Depicts a classification-based deep person re-id model, where $P^i$ refers to the $i^{th}$ person. (b) Illustrates a verification network whereby  the similarity $S$ and dissimilarity D for  a pair of images is found.  (c) A Triplet loss based DNN, where $A, P, N$ indicate anchor, positive, and negative samples, respectively. (d) A quadruplet based DNN 
	(e) Conventional diffusion-based DNN, which leverages the similarities among all the images in the gallery to learn a better similarity. (f) The proposed deep constrained dominant sets (DCDS), where, $\textcolor{red}{P}$ indicates the constraint (probe-image); and, images in the constrained cluster, the enclosed area, indicates the positive samples to the probe image.}
	
	\vspace{-0.6cm}
	\label{fig:metaphor}
\end{figure}

Person re-identification aims at retrieving the most similar images to the probe image, from a large scale gallery set captured by camera networks. Among the challenges which hinder person re-id tasks, include background clutter, Pose, view and illumination variation can be mentioned.

Person re-id can be taken as a person retrieval problem based on the ranked similarity score, which is obtained from the pairwise affinities between the probe and the dataset images. However, relying solely on the pairwise affinities of probe-gallery images, ignoring the underlying contextual information between the gallery images often leads to  an undesirable similarity ranking.  To tackle this,  several works  have been reported, which employ similarity diffusion to estimate a second order similarity that considers the intrinsic manifold structure of the given affinity matrix \cite{BaiBT17}, \cite{LoyLG13}, \cite{DonoserB13}, \cite{BaiZWBLT17}. Similarity diffusion is a process of exploiting the contextual information between all the gallery images to provide a  context sensitive similarity. Nevertheless, all these methods do not leverage the advantage of deep neural networks. Instead, they employ the similarity diffusion process as a post-processing step on the top of the DNN model. Aiming to improve the discriminative power of a DNN model, there have been recent works which incorporate a similarity diffusion process in an end-to-end manner \cite{ShenLXYCW18DGSRW},\cite{ShenLYCW18DSGNN},\cite{Chen0LSW18}. Following \cite{BertasiusTYS17}, which applies a random walk in an end-to-end fashion for solving semantic segmentation problem, authors in \cite{ShenLXYCW18DGSRW} proposed a group-shuffling random walk network for fully utilizing the affinity information between gallery images in both the training and testing phase. 
Also, the authors of \cite{ShenLYCW18DSGNN} proposed similarity-guided graph neural network (SGGNN) to exploit the relationship between several prob-gallery image similarities. 

However, most of the existing graph-based end-to-end learning methods apply the similarity diffusion without considering any constraint or attention mechanism to the specific query image. Due to  that the second order similarity  these methods yield is highly prone to noise. To tackle this problem, one possible mechanism could be to guide the similarity propagation by providing seed (or constraint) and let the optimization process estimate the optimal similarity between the seed and  nearest neighbors, while treating  the seed as our attention point. To formalize this idea, in this paper,  we model person re-id problem as finding an internally coherent and externally incoherent constrained cluster in an end-to-end fashion. To this end, we adopt a graph and game theoretic method called constrained dominant sets in an end-to-end manner. To the best of our knowledge, we are the first ones to integrate the well known unsupervised clustering method called dominant sets in a DNN model. To summarize, the contributions of the proposed work are:
\begin{itemize}
	\item  For the very first time, the  dominant sets clustering method is integrated in a DNN and optimized  in end-to-end fashion.
	\item A one-to-one correspondence between person re-identification and constrained clustering problem is established.
	\item  State-of-the-art results are significantly improved.
	
\end{itemize}

The paper is structured as follow. In section 2, we  review the related works. In section 3, we  discuss the proposed method with a brief introduction to dominant sets and constrained dominant sets. Finally, in section 4, we  provide an extensive experimental analysis on three different benchmark datasets.

\section{Related works}

Person re-id is one of the challenging computer vision tasks due to  the variation of illumination condition, backgrounds, pose and viewpoints. Most recent methods train DNN models with different learning objectives including verification, classification, and similarity learning \cite{ChengGZWZ16}, \cite{ZhaoLZW17}, \cite{VariorHW16}, \cite{AhmedJM15}. For instance, verification network (V-Net) \cite{LiZXW14VFn}, Figure \ref{fig:metaphor}(b), applies a binary classification of image-pair representation which is trained under the supervision of binary softmax loss. Learning accurate similarity and robust feature embedding has a vital role in the course of person re-identification process. Methods which integrate siamese network with contrastive loss are a typical showcase of deep similarity learning for person re-id \cite{ChenCZH17}. The optimization goal of these models is to estimate the minimum distance between the same person images, while maximizing the distance between images of different persons. However, these methods focus on the pairwise distance ignoring the contextual or relative distances. Different schemes have tried to overcome these shortcomings. In Figure \ref{fig:metaphor}(c),  triplet loss is exploited to enforce the correct order of relative distances among image triplets \cite{ChengGZWZ16}, \cite{DingLWC15}, \cite{ZhaoLZW17} . In Figure \ref{fig:metaphor}(d), Quadruplet loss \cite{ChenCZH17} leverages the advantage of both contrastive and triplet loss, thus it is able to maximize the intra-class similarity while minimizing the inter-class similarity. Emphasizing the fact that these methods entirely neglect the global structure of the embedding space, \cite{Chen0LSW18}, \cite{ShenLXYCW18DGSRW}, \cite{ShenLYCW18DSGNN} proposed graph based end-to-end diffusion methods shown in Figure \ref{fig:metaphor}(e).

\textbf{Graph based end-to-end learning.} Graph-based methods have played a vital role in the rapid growth of computer vision applications in the past.   However, lately, the advent of deep convolutional neural networks and their tremendous achievements in the field has attracted great attention of researchers. Accordingly, researchers have made a significant effort to integrate, classical methods, in particular, graph theoretical methods, in end-to-end learning. Shen \et\cite{ShenLYCW18DSGNN} developed two constructions of deep convolutional networks on a graph, the first one is based upon hierarchical clustering of the domain, and the other one is based on the spectrum of graph Laplacian. Yan \et\cite{YanXL18} proposed a model of dynamic skeletons called Spatial-Temporal Graph Convolutional Networks (ST-GCN), which provides a capability to automatically learn both the spatial and temporal pattern of data. Bertasius \et \cite{BertasiusTYS17} designed a convolutional random walk (RWN), where by jointly optimizing the objective of pixelwise affinity and semantic segmentation they are able to address the problem of blobby boundary and spatially fragmented predictions. Likewise, \cite{ShenLXYCW18DGSRW} integrates random walk method in end-to-end learning to tackle person re-identification problem. In \cite{ShenLXYCW18DGSRW}, through the proposed deep random walk and the complementary feature grouping and group shuffling scheme, the authors  demonstrate that one can estimate a robust probe-gallery affinity. Unlike recent Graph neural network (GNN) methods \cite{ShenLYCW18DSGNN}, \cite{KipfW16}, \cite{ShenLXYCW18DGSRW}, \cite{Chen0LSW18}, Shen \et \cite{ShenLYCW18DSGNN} learn the edge weights by exploiting the training label supervision,  thus they are able to learn more accurate feature fusion weights for updating nodes feature.

\textbf{Recent applications of dominant sets.} Dominant sets (DS) clustering \cite{PavPel07} and its constraint variant constrained dominant sets (CDS) \cite{ZemenePECCV16} have been employed in several recent computer vision applications ranging from person tracking \cite{TesfayeZPP16}, \cite{TesfayeZPPS17}, geo-localization \cite{ZemeneTIPPS19}, image retrieval \cite{ZemeneAP16}, \cite{AlemuPelillo}, 3D object recognition \cite{WangPS17}, to Image segmentation and co-segmentation \cite{ZemeneAP17}. Zemene \et \cite{ZemenePECCV16}  presented CDS with its applications to interactive Image segmentation. Following, \cite{ZemeneAP17} uses CDS to tackle both image segmentation and co-segmentation in interactive and unsupervised setup. Wang \et \cite{WangPS17} recently used dominant sets clustering in a recursive manner to select representative images from a collection of images and applied a pooling operation on the refined images,  which survive at the recursive selection process. Nevertheless, {\em none of the above works have attempted to leverage the dominant sets algorithm in an end-to-end manner.}

In this work, unlike most of the existing graph-based DNN model, we propose a constrained clustering based scheme in an end-to-end fashion,  thereby,  leveraging the contextual information hidden in the relationship among person images. In addition, the proposed scheme significantly magnifies the inter-class variation of different person-images while reducing the intra-class variation of the same person-images. The big picture of our proposed method is depicted in Figure \ref{fig:metaphor}(f), as can be seen, the objective here is to find a coherent constrained cluster which incorporates the given probe image $P$.
\vspace{-0.1cm}
\section{Our Approach}
\vspace{-0.1cm}
In this work, we cast probe-gallery matching as optimizing a constrained clustering problem, where the probe image is treated as a constraint, while the positive images to the probe are taken as members of the constrained-cluster. Thereby, we integrate such clustering mechanism into a deep CNN to learn a robust  features through the leveraged contextual information. This is achieved by traversing through the global structure of the given graph to induce a compact set of images based on the given initial similarity(edge-weight).
\begin{figure*}[t]
	
	\vspace{-0.6cm}
	\begin{center}
		
		\includegraphics[width=0.8\linewidth ,trim=0cm 0cm 0cm 0cm,clip]{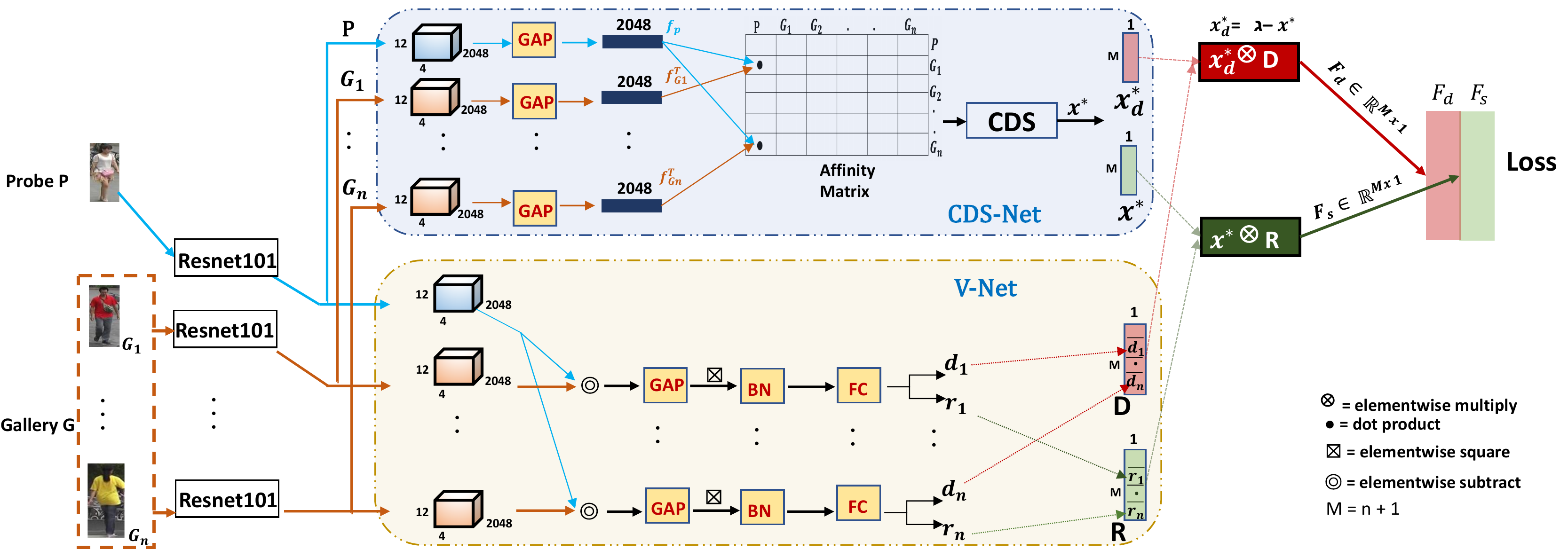}
	\end{center}
	\vspace{-0.3cm}
	\caption{Workflow of the proposed DCDS. Given n number of gallery images, $G,$ and probe image $P$, we first extract their Resent101 features right before the global average pooling (GAP) layer, which are then fed to CDS-Net (upper stream)  and V-Net (lower stream) branches. 
	In the CDS-branch, after applying GAP, we compute the similarity between $M^2$ pair of probe-gallery image features, $f_p$ and $f_{Gi}^T$ using their dot products, where $T$ denotes a transpose. Thereby, we obtain $M \times M$ affinity matrix. Then, we run CDS taking the probe image as a constraint to find the solution $x^* \in {\rm I\!R}^{M \times 1}$ (similarity), and the dissimilarity, $x^*_d,$ is computed as an additive inverse of the similarity $x^*.$ Likewise, in the lower stream we apply elementwise subtraction on $M$ pair of probe-gallery features.  This is followed by GAP, batch normalization (BN), and fully connected layer (FC) to obtain probe-gallery similarity score, $R \in {\rm I\!R}^{M \times 1},$ and probe-gallery dissimilarity score, $D \in {\rm I\!R}^{M \times 1}$. Afterward, we  elementwise multiply $x^*$  and $R,$ and $x^*_d$ and $D$, to find the final similarity, $F_s,$ and disimilarity, $F_d,$ scores, respectively. Finally, to find the prediction loss of our model, we apply a cross entropy loss, the ground truth ($G_t$) is given as $G_t  \in {\rm I\!R}^{M \times 1}$. }
	\label{fig:pipline}
\end{figure*}

\subsection{Dominant Sets and Constrained Dominant Sets} Dominant sets is a graph theoretic notion of a cluster, which generalizes the concept of a maximal clique to edge-weighted graphs.  First, the data to be clustered are represented as an undirected edge-weighted graph with no self-loops, $G = (V, E,w)$, where $V = \{1, . . . , M\}$ is the vertex set, $E \subseteq V \times V$ is the edge set, and $w : E \rightarrow R_+^*$ is the (positive) weight function. Vertices in $G$ correspond to data points, edges represent neighborhood relationships, and edge-weights reflect similarity between pairs of linked vertices. As customary, we represent the graph $G$ with the corresponding weighted adjacency (or similarity) matrix, which is the $M \times M$ nonnegative, symmetric matrix $A = (a_{ij})$, defined as $a_{ij} = w(i, j)$, if $(i, j) \in E$, and $a_{ij} = 0$ otherwise. Note that the diagonal elements of the adjacency matrix A are always set to zero indicating that there is no self-loops in graph $G.$ As proved in \cite{PavPel07}, one can extract a coherent cluster from a given graph by solving a quadratic program $f(\x)$ as,
\vspace{-0.4cm}
\begin{equation}
\label{eq2}
\begin{array}{ll}
\text{maximize }  &  f(\x) = \x' A \x, \\
\text{subject to} &  \mathbf{x} \in \Delta
\end{array}
\end{equation}

where, $\Delta$ is the standard simplex of $R^n$.
Zemene et. al \cite{ZemenePECCV16} proposed an extension of dominant sets which allows one to constrained the clustering process to contain intended constraint nodes $P$. Constrained dominant set (CDS) is an  extensions of dominant set which contains a parameterized regularization term that controls the global shape of the energy landscape. When the regularization parameter is zero the local solutions are known to be in one-to-one correspondence with the dominant sets. A compact constrained cluster could be easily obtained from a given graph by defining a paramertized quadratic program as,
\vspace{-0.1cm}
\begin{equation}
\label{eqn:parQP}
\begin{array}{ll}
\text{maximize }  &  f_P^\alpha(X) = \x' (A - \alpha \hat I_P) \x, \\
\text{subject to} &  \mathbf{x} \in \Delta
\end{array}
\end{equation}
where, $\hat I_P$ refers to $M \times M$ diagonal matrix whose diagonal elements are set to zero in correspondence to the probe $P$ and to 1 otherwise. Let $\alpha > \lambda_{max}(A_{V \backslash P} ),$ where $\lambda_{max}(A_{V\backslash P})$ is the largest eigenvalue of the principal submatrix of $A$ indexed by the element of $V\backslash P.$ If $\x$ is a local maximizer of $ f_P^\alpha(\x)$ in $\Delta,$ then $\delta(\x) \cap P \neq \emptyset,$ where, $\delta(\x) = {i \in V : \x_i > 0.}$ We refer the reader to \cite{ZemenePECCV16} for the proof.
Equations \ref{eq2} and \ref{eqn:parQP} can be simply solved with a straightforward continuous optimization technique from evolutionary  game theory called replicator dynamics, as follows:
\vspace{-0.2cm}
\begin{equation}
\label{eqn:Replicator}
x_i{(t+1)} = x_i{(t)} \frac{(A\x{(t)})_i}{\x{(t)}'A\x{(t)}}.
\vspace{-0.2cm}
\end{equation}
for $i$ = $1, . . . , M.$
\begin{figure*}[t]
	\vspace{-0.3cm}
	
	\begin{center}
		
		\includegraphics[width=0.92\linewidth ,trim=0cm 0cm 0cm 0cm,clip]{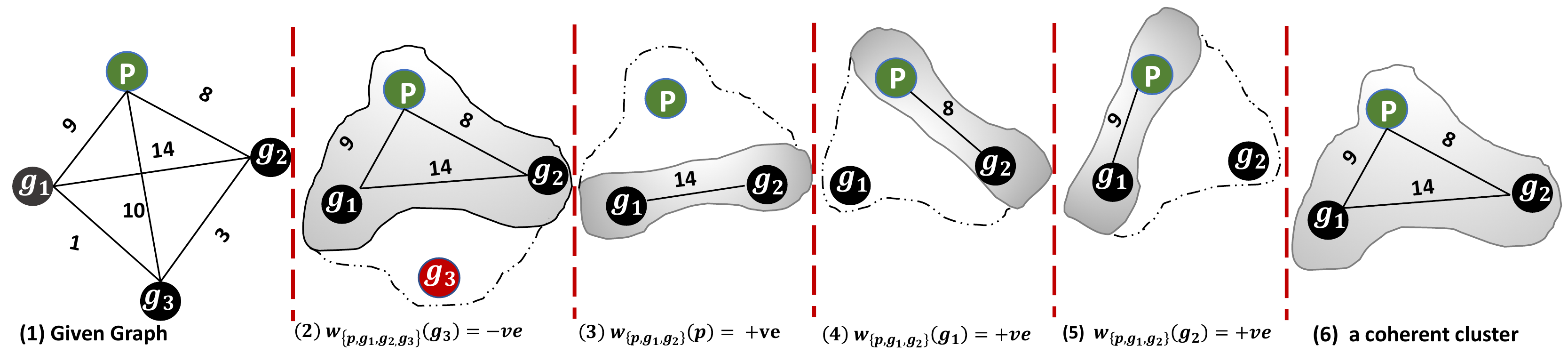}
	\end{center}
	\vspace{-0.5cm}
	\caption{Let $S= \{P,g_1, g_2, g_3\}$ comprises probe, $P,$ and gallery images $g_i$. As can be observed from the above toy example, the proposed method asses the contribution of each participant node $i \in S$ with respect to the subset $S \backslash i$. (1) shows graph G, showing the pairwise similarities of query-gallery images. (2-5) show the relative weight
	(Equ. \ref{eqn:wdegree}) of each node with respect to the overall similarity between $i$ and $S \backslash i$. (2) $w_{p,g_1,g_2, g_3}{(g_3)} < 0,$ shows that the Node $\{g_3 \}$ with Node $\{ P, g_1, g_2\}$ has a negative impact on the coherency of the cluster. (3) shows that clustering $\{P \}$ with $\{g_1 \}$ and $\{ g_2\}$ has a positive contribution to the compactness of set $\{P, g_1, g_2\}.$ (4), similarly, shows the relative weight of $g_1,$ $w_{p,g_1,g_2}{(g_1)} > 0.$ (5) shows the relative weight of $g_2, w_{p,g_1,g_2}{(g_2)} > 0$. And, (6) is a coherent subset (cluster) extracted from the graph given in (1).}
	\label{fig:exampler}
	\vspace{-0.2cm}
\end{figure*}

\subsection{Modeling person re-id as a Dominant Set}
Recent methods \cite{Chen0LSW18}, \cite{BertasiusTYS17} have proposed different models, which leverage local and group similarity of images in an end-to-end manner. Authors in \cite{Chen0LSW18} define a group similarity which emphasizes the advantages of estimating a similarity of two images, by employing the dependencies among the whole set of images in a given group. In this work, we establish a natural connection between finding a robust probe-gallery similarity and constrained dominant sets. Let us first elaborate the intuitive concept of finding a coherent subset from a given set based on the global similarity of given images. For simplicity, we represent  person-images as vertices of graph $G,$ and their similarity as edge-weight $w_{ij}$. Given vertices $V,$ and $S \subseteq V$  be a non-empty subset of vertices and $i \in S$, average weighted degree of each  $i$ with regard to $S$ is given as $\vspace{-0.4cm}$  $$ \label{eq1} \phi_S(i,j)=a_{ij}-\frac{1}{|S|} \sum_{k \in S} a_{ik}~,
\vspace{-0.2cm}$$ 
where $\phi_S(i,j)$ measures the (relative) similarity between node $j$ and $i$, with respect to the average similarity between node $i$ and its neighbors in $S$. Note that $\phi_S(i,j)$ can be either positive or negative. Next, to each vertex $i \in S$ we assign a weight defined (recursively) as follows:
\vspace{-0.2cm}
\begin{equation}
\vspace{-0cm}
\label{eqn:wdegree}
w_S(i)=
\begin{cases}
1,&\text{if\quad $|S|=1$},\\
\sum_{j \in S \setminus \{i\}} \phi_{S \setminus \{i\}}(j,i)w_{S \setminus \{i\}}(j),&\text{otherwise}
\end{cases}
\end{equation}
where $w_{ij} (i) = w_{ij} (j) = a_{ij}$  for all $i, j  \in V( i \neq j)$. \\
Intuitively, $w_S(i)$ gives us a measure of the overall similarity between vertex $i$ and the vertices of $S\setminus \{i\}$, with respect to the overall similarity among the vertices in $S\setminus \{i\}$. Hence, a \textbf{positive} $w_S(i)$ indicates that adding $i$ into its neighbors in $S$ will raise the internal coherence of the set, whereas in the presence of a \textbf{negative} value we expect the overall coherence to  decline. In CDS, besides the additional feature, which allows us to incorporate a constraint element in the resulting cluster, all the characters of DS are inherited.
\vspace{-0.4cm}
\subsubsection{A Set of a person images as a constrained cluster}
We cast person re-identification as finding a constrained cluster, where, elements of the cluster correspond to a set of same person images and the constraint refers to the probe image used to extract the corresponding cluster. As customary, let us consider  a given mini-batch with $M$ number of person-images, and each mini batch with $k$  person identities (ID), thus, each person-ID has $\Omega = M / k$ images in the given mini-batch. Note that, here, instead of a random sampling we design a custom sampler which samples $k$ number of person IDs in each mini-batch. Let $B = \{I_{p_1}^1, . . . I_{p_1}^{\Omega}, I_{p_2}^1, . . . I_{p_2}^{\Omega}, . . . I_{p_k}^1, . . . I_{pk}^{\Omega} \}$ refers to the set of images in a single mini-batch.  Each time when we consider image $I_{p_1}^1$ as a probe image $P$, images which belong to the same person id, $\{I_{p_1}^2, I_{p_1}^3 . . . I_{p_1}^k \},$ should be assigned a large membership score to be in that cluster. In contrast, the remaining images in the mini-batch should be assigned significantly smaller membership-score to be part of that cluster.
Note that our ultimate goal here is to find a constrained cluster which comprises all the images of the corresponding person given in that specific mini-batch. Thus, each participant in a given mini-batch is assigned a membership-score to be part of a cluster. Furthermore, the characteristics vector, which contains the membership scores of all participants is always a stochastic vector, meaning that  $\sum_{i =1 }^M z_i = 1,$ where $z_i$ denotes the membership score of each image in the cluster.

As can be seen from the toy example in Figure \ref{fig:exampler}, the initial pairwise similarities between the query and gallery images hold  valuable information, which define the relation of nodes in the given graph. However, it is not straightforward to redefine the initial pairwise similarities in a way which exploit the inter-images relationship. Dominant Sets (DS) overcome this problem with defining a  weight of each image $p, g_1, g_2, g_3$ with regard to subset $S\backslash i$ as depicted in Figure$\ref{fig:exampler}, (2-5),$ respectively. As can be observed from Figure \ref{fig:exampler}, adding node $g_3$ to cluster $S$  degrades the coherency of cluster $S = \{p, g_1, g_2, g_3\},$ whereas the relative similarity of the remaining images with respect to set $S = \{p, g_1, g_2\}$ has a positive impact on the coherency of the cluster. 
It is evident that the illustration in Figure \ref{fig:exampler} verifies that the proposed DCDS (Deep Constrained Dominant Set) could easily measure the contribution of each node in the graph and utilize it in an end-to-end learning process. Thereby, unlike a siamese, triplet and quadruplet based contrastive methods, DCDS consider the whole set of images in the mini-batch to measure the similarity of image pairs and enhance the learning process.

\subsection{CDS Based End-to-end Learning}
In this section, we discuss the integration of CDS in end-to-end learning. We adopt a siamese based Resent101, with a novel  verification loss to find probe-gallery similarity, $R$, and dissimilarity, $D$ scores. 
As can be seen from Figure \ref{fig:pipline}, we  have two main branches:  CDS network branch (CDS-Net) and verification network branch (V-Net). 
In the CDS-Net, the elements of pairwise affinity matrix are computed first as a dot product of the global pooling feature of a pair of images. Afterward, the replicator dynamics \cite{Wei95} is applied, which is a discrete time solver of the parametrized quadratic program, Equ. \ref{eqn:parQP2}, whose solution corresponds to the CDS. Thus, assuming that  there are $M$ images in the given mini-batch, the replicator dynamics, Equ. \ref{eqn:Replicator}, is  recursively applied $M$ times taking each image in the mini-batch as a constraint.  Given graph $G = (V, E,w)$ and its corresponding adjacency matrix $A \in R^{M \times M},$ and probe $P \subseteq V.$ First, a proper modification of the affinity matrix $A$ is applied by setting parameter $-\alpha$ to the diagonal corresponding to the subset $V \backslash P$ and zero to the diagonal corresponding to the constraint image $P$. 
Next, the modified adjacency matrix, $B,$ is feed to the Replicator dynamics, by initiating the dynamics with a characteristic vector of uniform distribution $x^{t_0}$, such  that  initially all the images in the mini-batch are assigned equal membership probability to be part of the cluster. Then, to find a constrained cluster  a parametrized quadratic program is defined as:
\begin{equation}
\label{eqn:parQP2}
\begin{array}{ll}
\text{maximize }  &  f_P^\alpha(\x)^i = \x' B \x \quad  where,  B = A - \alpha \hat I_p. \\ 
\text{subject to} &  \mathbf{x} \in \Delta
\end{array}
\end{equation}

The solution, $ \x_i^*,$ of $ f_P^\alpha(\x)^i$  is a characteristics vector which indicates the probability of each gallery image to be included in a cluster, containing the probe image $P^i$. Thus, once we obtain the CDS, $\x_i^* = [z^i_{g_1}, z^i_{g_2} . . . z^i_{g_M}],$  for each probe image, we store each solution $ \x_i^*$, in $Y \in {\rm I\!R}^{M \times M},$ as 
$$
Y = 
\begin{pmatrix} 
 \x_i^* \\ 
\vdots \\
\x_M^*\\ 
\end{pmatrix} =
\begin{pmatrix} 
z_{g_1}^{1 }&z_{g_2}^{1 }& \cdots  & z_{g_M}^{1} \\ 
\vdots && ~~ \ddots ~~ & \vdots &\\
z_{g_1}^{M }&z_{g_2}^{M }& \cdots  & z_{g_M}^{M}\\ 
\end{pmatrix}.
$$ 
Likewise, for each probe, $P^i,$ we store the probe-gallery similarity, R, and dissimilarity, D, obtained from the V-Net (shown in Figure \ref{fig:pipline}) in $S'$ and $D'$ as, $S' = [R^1, R^2, . . . R^M]$ and $D' = [D^1, D^2, . . . D^M].$
Next, we fuse the similarity obtained from the CDS branch with the similarity  from the V-Net as 
\begin{equation}
\label{eqn:fus}
\begin{array}{ll}
F_s = \beta(Y) \otimes (1 - \beta)(S'),  \\ 
F_d = \beta(Y_d) \otimes (1 - \beta)(D'), \quad where, \quad Y_d = \delta - Y
\end{array}
\end{equation}
$\delta$ is empirically set to 0.3. We then vectorize $F_s$ and $F_d$ into ${\rm I\!R}^{(M^2 \times 2)},$ where, the first column stores the dissimilarity score, while the second column stores the similarity score. Afterward, we simply apply cross entropy loss to find the prediction loss.  
The intriguing feature of our model is that it does not need any custom optimization technique, it can be end-to-end optimized through a standard back-propagation algorithm. Note that, Figure \ref{fig:pipline} illustrates the case of a single probe-gallery, whereas Equ. \ref{eqn:fus} shows the solution of $M$ probe images in a given mini-batch.
\begin{figure}[t]

	\begin{center}
		
		\includegraphics[width=1\linewidth ,trim=0cm 0cm 0cm 0cm,clip]{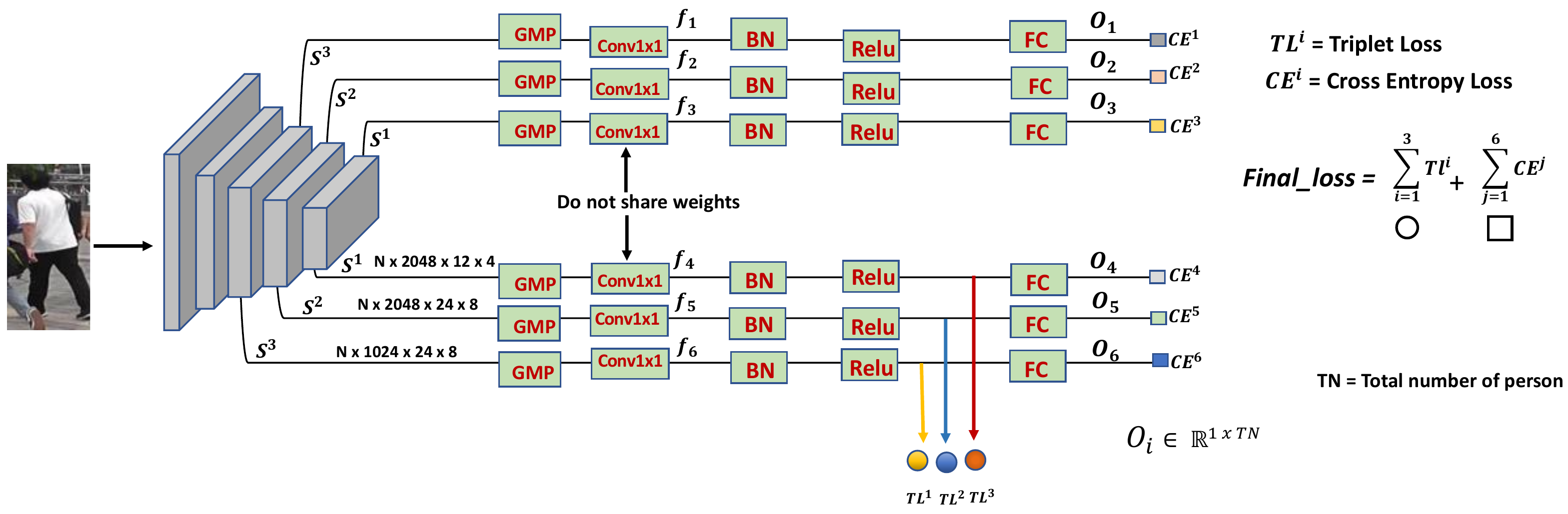}
	\end{center}
	\vspace{-0.3cm}
	
	\caption{Illustrates the auxiliary net, which consists of two branches which are jointly trained. We first  use  features at  different layers, $S_1, S_2, S_3,$ and then feed these to Global Maxpooling (GMP), Conv, BN, Relu and FC layers for further encoding. 
	We then compute triplet losses employing the features from the lower three streams after Relu, shown  by yellow, blue, and red circles. Next, after the final FC layer, we compute the cross-entropy loss for each of the six different outputs, $O_i,$ from the upper and lower stream  shown by distinct colored-boxes. Note that even if the upper and lower stream apply the same operations, on $S_1, S_2$ and $S_3,$ they do not share the weights; thus the encoding is different. We finally compute the final loss as the sum of the average of the triplet and cross entropy losses.}
	\label{fig:Auxnet}
\end{figure}

\subsection{Auxiliary Net}

\begin{figure}
	
	\vspace{-0.4cm}
	\begin{center}
		
		\includegraphics[width=1\linewidth ,trim=0cm 0cm 0cm 0cm,clip]{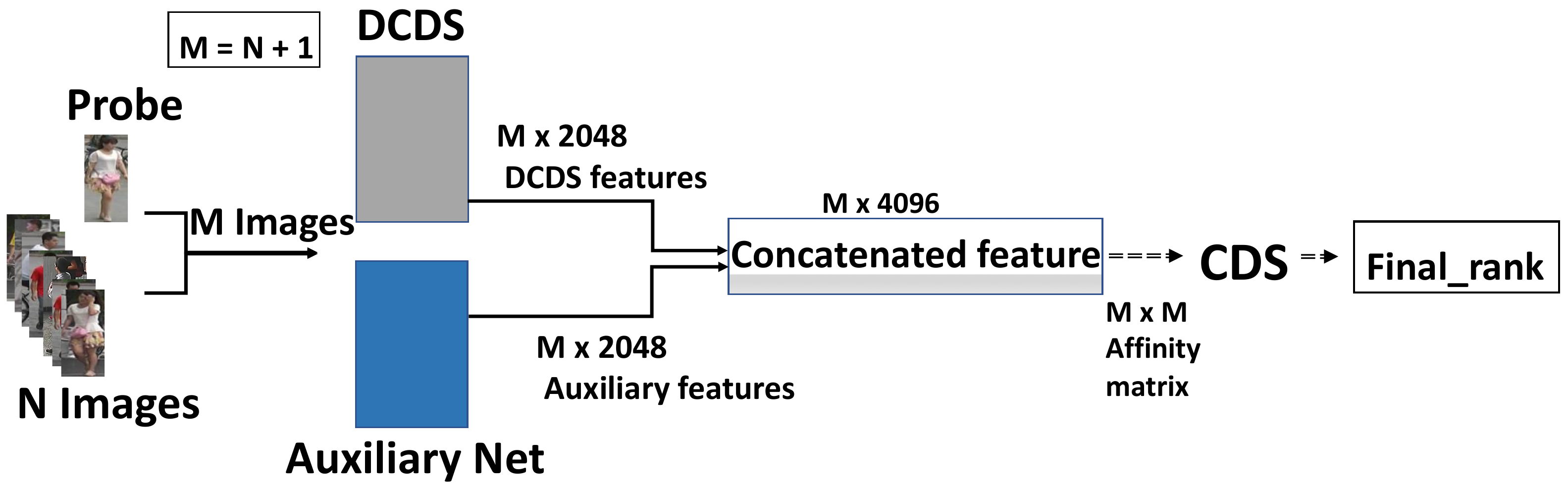}
	\end{center}
	\vspace{-0.3cm}
	\caption{During testing, given a probe and gallery images, we extract DCDS and auxiliary features and concatenate them to find a single vector. Afterward, we build M x M affinity matrix and run CDS with constraint expansion mechanism to find the final probe-gallery similarity rank.}
	\label{fig:Testphase}
	\vspace{-0.5cm}
\end{figure}

In this work, we integrate an auxiliary net to further improve the performance of our model. The auxiliary net is trained based on the multi-scale prediction of Resnet50 \cite{HeZRS16}. It is a simple yet effective architecture, whereby we can easily  compute both triplet and cross entropy loss of different layers of Resnet50 \cite{HeZRS16}, hence further enhancing the learning capability. Consequently, we compute the average of both losses to find the final loss. As can be observed from Figure \ref{fig:Auxnet}, we employ three features at different layers from Resnet50 $conv5\_x$ Layer, and then we fed these three features to the subsequent layers, MP, Conv, BN, and FC layers. Next, we compute triplet and cross entropy loss for each feature which comes from the Relu and FC layers, respectively. During testing phase we concatenate the features that come from the DCDS and Auxiliary Net to find 4096 dimensional feature. We then apply CDS to find the final ranking$\_$score, (See Figure \ref{fig:Testphase}).

\vspace{-0.1cm}
\subsection{Constraint Expansion During Testing}
\begin{figure}[t]

	\begin{center}
		
		\includegraphics[width=1\linewidth ,trim=0cm 5.4cm 0cm 0cm,clip]{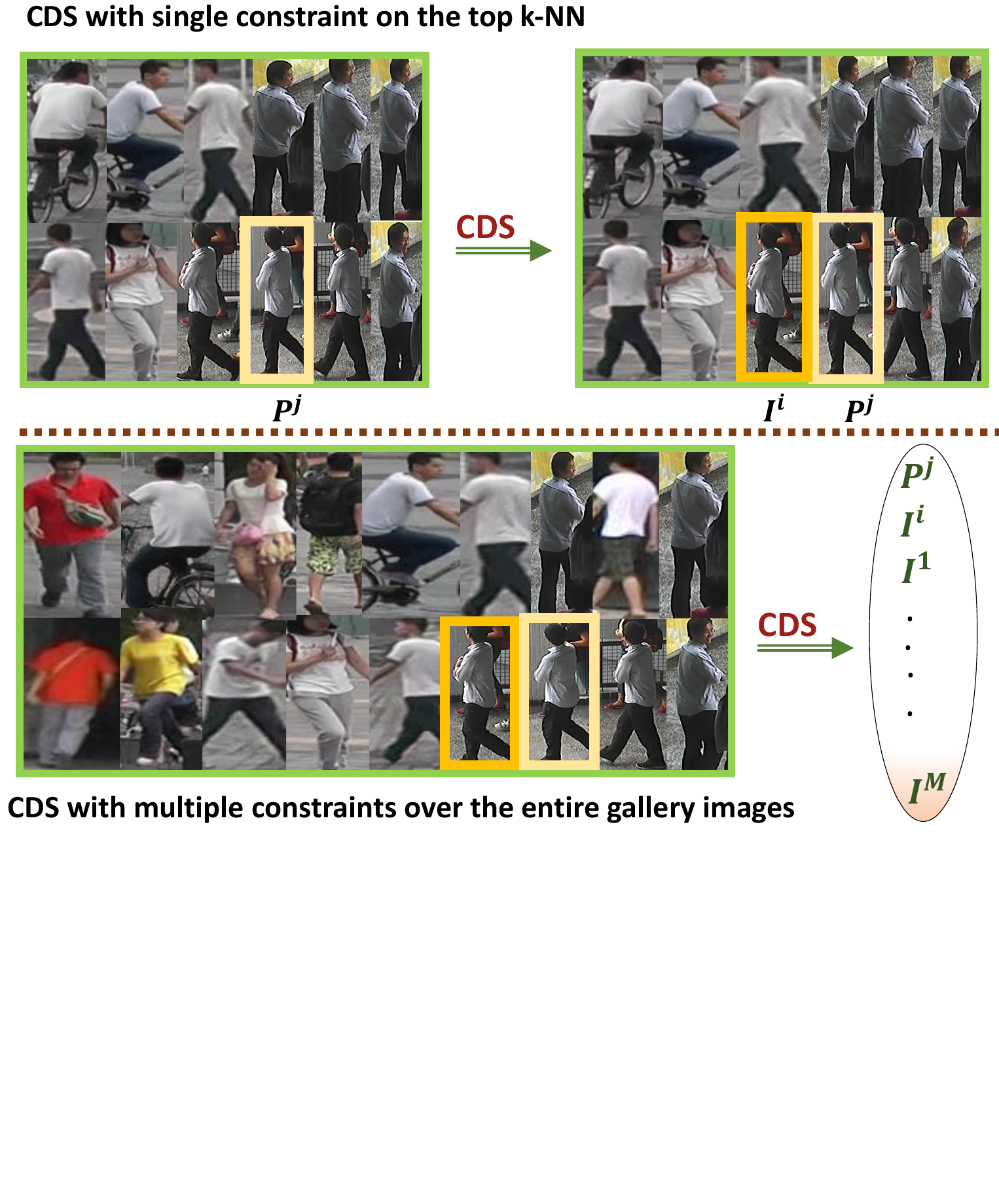}
	\end{center}
		\vspace{-0.3cm}
	\caption{Given a constraint (probe-image) $P^j,$ we first collect k-NNs to the probe-image, based on the pairwise similarities. Subsequently, we run CDS on the graph of the k-NN. Then, based on the cluster membership score obtained, we choose image $I^{i }$,  with the highest membership score and re-run CDS, considering $P^j$ and $I^i$ as constraints, over the graph of the all set of  images, $I^M,$ in the minibatch. Afterward, we consider the solution as our final rank.}
	\label{fig:Conexpansion}
	\vspace{-0.1cm}
\end{figure}

We propose a new scheme (illustrated in Figure \ref{fig:Conexpansion}) to expand the number of constraints in order to guide the similarity propagation during the testing phase. Given an affinity matrix, which is constructed using the features obtained from the concatinated feature (shown in Figure \ref{fig:Testphase}), we first collect k-NN's of the probe image. Then, we run CDS on the graph of the NN’s. Next, from the resulting constrained cluster, we select the one with the highest membership score, which is used as a constraint in the subsequent step. We then use multiple-constraints and run CDS over the entire graph.

\section{Experiments}
To validate the performance of our method we have conducted several experiments on three publicly available benchmark datasets, namely CUHK03 \cite{LiZXW14VFn}, Market1501 \cite{ZhengSTWWT15}, and DukeMTMC-reID \cite{ZhengZY17}.

\subsection{Datasets and evaluation metrics}
\textbf{Datasets:} CUHK03 \cite{LiZXW14VFn} dataset comprises 14,097 manually and automatically cropped images of 1,467 identities, which are captured by two cameras on campus; in our experiments, we have used manually annotated images. Market1501 dataset \cite{ZhengSTWWT15} contains 32,668 images which are split into 12, 936 and 19,732 images as training and testing set, respectively. Market1501 dataset has totally 1501 identities which are captured by five high-resolution and one low-resolution cameras, the training and testing sets have 751 and 750 identities respectively. To obtain the person bounding boxes, Deformable part Model (DPM) \cite{FelzenszwalbGMR10} is utilized. DukeMTMC-reID  is generated from a tracking dataset called DukeMTMC. DukeMTMC is captured by 8 high-resolution cameras, and person-bounding box is manually cropped; it is organized as 16,522 images of 702 person for training and 18, 363 images of 702 person for testing.\\
\textbf{Evaluation Metrics:} Following the recent person re-id methods, we  use mean average precision (mAP) as suggested in \cite{ZhengSTWWT15}, and Cumulated Matching Characteristics (CMC) curves to evaluate the performance of our model. Furthermore, all the experiments are conducted using the standard single query setting \cite{ZhengSTWWT15}. 

\begin{table}[h]
	
	\begin{center}
		
		\begin{tabular}{l|c|c|c} 
			\hline
			Methods & mAP& rank-1 & rank-5\\
			
			\hline\hline
			SGGNN \cite{ShenLYCW18DSGNN} ECCV18& 82.8&92.3& 96.1    \\
			DKPM \cite{ShenXLYW18DKPM} CVPR18&75.3&90.1& 96.7    \\
			DGSRW \cite{ShenLXYCW18DGSRW} CVPR18& 82.5&92.7& 96.9    \\
			GCSL \cite{Chen0LSW18} CVPR18&  81.6&93.5& - \\  
			CPC \cite{WangCW018} CVPR18& 69.48&83.7& -    \\		  
			MLFN \cite{ChangHX18} CVPR18&  74.3&90.0& - \\   
			HA-CNN \cite{LiZG18} CVPR18& 75.7 &91.2& -   \\		 
			PA \cite{SuhWTML18} ECCV18&  74.5& 88.8& 95.6 \\ 
			HSP \cite{KalayehBGKS18} CVPR18&  83.3 & 93.6& 97.5 \\
			{\bf Ours}  & \textbf{85.8}& \textbf{94.81}& \textbf{98.1} \\
			\hline\hline
			$RA_{w/ RR}$ \cite{WangWYZCLHHW18} CVPR18& 86.7&90.9& -    \\	
			$PA _{w/ RR}$ \cite{SuhWTML18} ECCV18&  89.9& 93.4& 96.4 \\ 
			$HSP_{w/ RR}$ \cite{KalayehBGKS18} CVPR18 &  90.9& 94.6& 96.8 \\	
			{\bf Ours$_{w/ RR}$} & \textbf{93.3}& \textbf{95.4}&  \textbf{98.3} \\
			
			\hline
		\end{tabular}
	
	\end{center}
	
	\caption{A comparison of the proposed method with state-of-the-art methods on Market1501 dataset. Upper block, without re-ranking methods. Lower block, with re-ranking method, $w/ RR$, \cite{ZhongZCL17}.}
	\label{table:market}
	\vspace{-0.5cm}
\end{table}
\vspace{-0.1cm}

\subsection{Implementation Details}
We implement DCDS based on Resnet101 \cite{HeZRS16} architecture, which is pretrained on imagenet dataset. We adopt the training strategy of Kalayeh \et \cite{KalayehBGKS18}, and aggregate eight different person re-id benchmark dataset to train our model. In total, the merged dataset contains 89,091 images, which comprises 4937 person-ID (detail of the eight datasets is given in the supplementary material). We first train our model using the merged dataset (denoted as multi-dataset (MD)) for 150 epochs and fine-tune it with CUHK03, Market1501, and DukeMTMC-reID dataset. To train our model using the merged dataset, we set image resolution to 450 $\times$ 150. Subsequently, for fine-tuning the model we set image resolution to 384 $\times$ 128. Mini-batch size is set to 64, each mini-batch has 16 person-ID and each person-ID has 4 images. We also experiment only using a single dataset for training and testing, denoted as single-dataset (SD). For data augmentation, we apply random horizontal flipping and random erasing \cite{Zhong}. For optimization we use Adam, we initially set the learning rate to 0.0001, and drop it by 0.1 in every 40 epochs. The fusing parameter in Equ. \ref{eqn:fus}, $\beta$, is set to 0.9. 

\begin{table*}[h]
	
	\begin{center}
		
		\begin{tabular}{l|c|c|c|c|c|c|c|c} 
			\hline
			 \multirow{2}{*}{Methods}&\multicolumn{2}{c}{Market1501}&&\multicolumn{1}{c}{CUHK03}&&\multicolumn{2}{c}{DukeMTMC-reID}\\
			
		   & mAP& rank-1 & rank-5&rank-1&rank-5&mAP& rank-1&rank-5\\
			
			\hline\hline
			
			Baseline SD & 72.2& 86.5&  94.0&87.1&94.3&61.1&77.6&87.3 \\
			Baseline MD & 74.3& 87.5&  95.3 &87.7&95.2&62.3&79.1&88.8\\
			DCDS (SD ) & 81.4& 93.3&  97.6 &93.1&98.8&69.1&83.3&89.0\\
			DCDS (MD) & 82.3& 93.7&  98.0 &93.9&98.9&70.5&84.0&90.3\\
			Ours (SD + Auxil Net) & 83.0& 93.9&  98.2 &95.4&99.0&74.4&85.6&93.7\\
			{\bf Ours (MD + Auxil Net)} & {\bf 85.8}& {\bf 94.1}&  {\bf 98.1} &{\bf 95.8} & {\bf 99.1}& {\bf 75.5}& {\bf 86.1} & {\bf 93.2}\\
			
			\hline
		\end{tabular}
	\end{center}
	
	\caption{Ablation studies on the proposed method. SD and MD respectively refer to the method  trained on single and multiple-aggregated datasets. Baseline is the proposed method without CDS branch.}
	\label{table:ABLATION}
	\vspace{-0.3cm}
\end{table*}

\begin{table}
	
	\begin{center}
		
		\begin{tabular}{l|c|c} 
			\hline
			Methods & rank-1 & rank-5\\
			
			\hline\hline
			SGGNN \cite{ShenLYCW18DSGNN} ECCV18& 95.3 &99.1   \\
			DKPM \cite{ShenXLYW18DKPM} CVPR18& 91.1 & 98.3   \\
			DGSRW \cite{ShenLXYCW18DGSRW} CVPR18& 94.9 & 98.7   \\
			GCSL \cite{Chen0LSW18} CVPR18& 90.2& 98.5\\  
			MLFN \cite{ChangHX18} CVPR18&  89.2&- \\ 
			CPC \cite{WangCW018} CVPR18& 88.1 & -   \\		  
			PA \cite{SuhWTML18} ECCV18 & 88.0& 97.6 \\ 
			HSP \cite{KalayehBGKS18} CVPR18& 94.28& 99.04 \\
			{\bf Ours}  & {\bf 95.8} &  {\bf 99.1} \\
			\hline
		\end{tabular}
	\end{center}
	
	\caption{A comparison of the proposed method with state-of-the-art methods on CUHK03 dataset.}
	\label{table:compCUHK}
	\vspace{-0.2cm}
\end{table}

\begin{table}[h]
	
	\begin{center}
		
		\begin{tabular}{l|c|c|c} 
			\hline
			Methods & mAP& rank-1 & rank-5\\
			
			\hline\hline
			SGGNN \cite{ShenLYCW18DSGNN} ECCV18& 68.2 &81.1& 88.4  \\
			DKPM \cite{ShenXLYW18DKPM} CVPR18& 63.2 &80.3& 89.5   \\
			DGSRW \cite{ShenLXYCW18DGSRW} CVPR18& 66.4&80.7& 88.5    \\
			GCSL \cite{Chen0LSW18} CVPR18&  69.5&84.9& - \\  
			CPC \cite{WangCW018} CVPR18& 59.49 &76.44& -  \\		  
			MLFN \cite{ChangHX18} CVPR18&  62.8&81.0& - \\   
			RAPR \cite{WangWYZCLHHW18} CVPR18& 80.0&84.4& -    \\	
			PA \cite{SuhWTML18} ECCV18&  64.2& 82.1& 90.2 \\ 
			HSP \cite{KalayehBGKS18} CVPR18 &  73.3& 85.9& 92.9 \\
			Ours & \textbf{75.5} & \textbf{87.5}&  - \\
			\hline\hline
			$PA _{w/ RR}$ \cite{SuhWTML18} ECCV18&  83.9 & 88.3& 93.1\\ 
			$HSP_{w/ RR}$ \cite{KalayehBGKS18} CVPR18 &  84.99& 88.9& 94.27 \\
			{\bf Ours} $_{w/ RR}$ & \textbf{86.1}& 88.5&  - \\
			
			\hline
		\end{tabular}
	\end{center}
	
	\caption{A comparison of the proposed method with state-of-the-art methods on DukeMTMC-reID dataset.Upper block, without re-ranking methods. Lower block, with re-ranking method,$w/ RR$, \cite{ZhongZCL17}.}
	\label{table:compDuke}
	\vspace{-0.4cm}
\end{table}

\begin{figure}[t]
	\vspace{-0.1cm}
	
	\begin{center}
		
		\includegraphics[width=1\linewidth ,trim=0cm 0cm 0cm 0cm,clip]{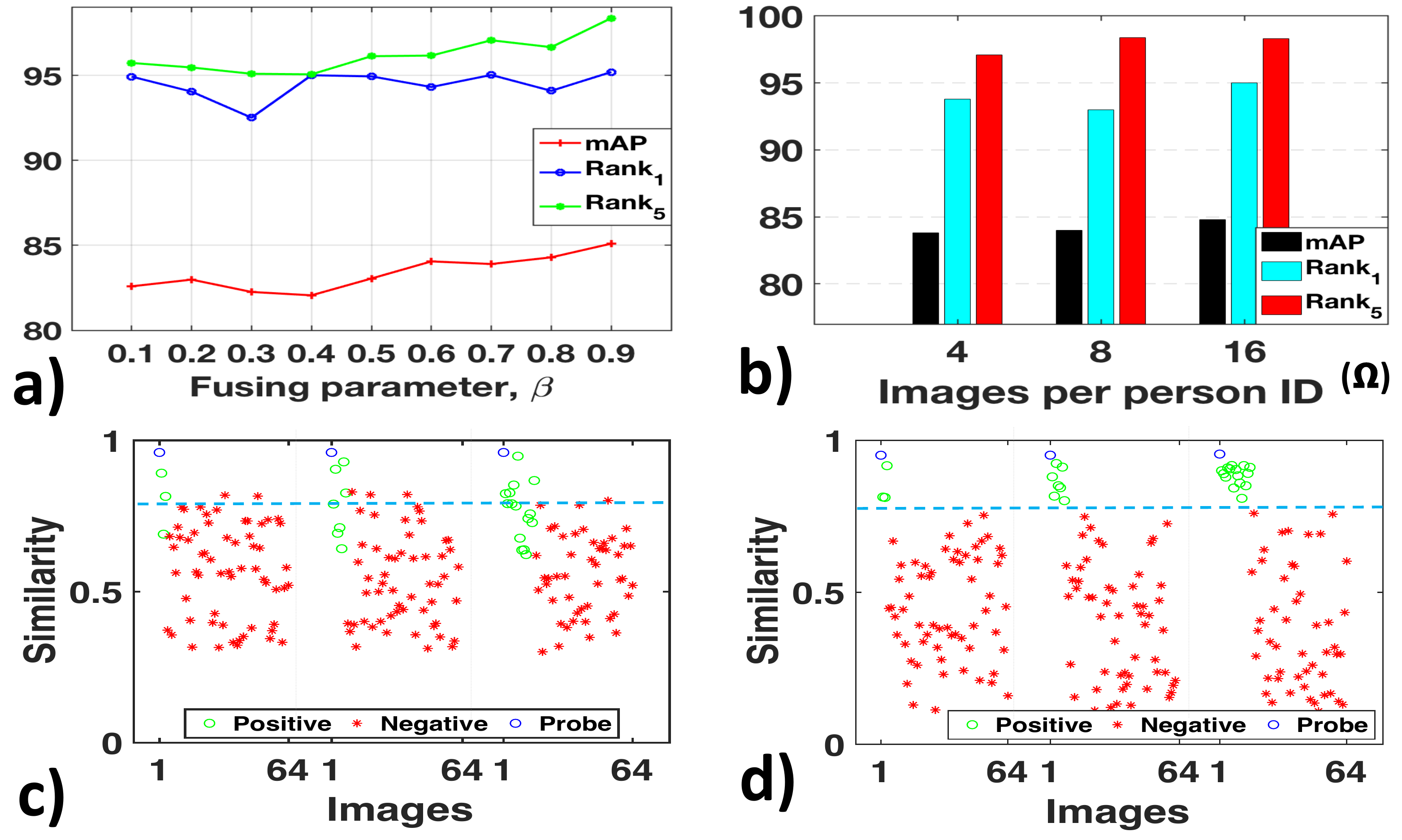}
	\end{center}
	
	\caption{Illustrates different experimental analysis performed on Market1501 dataset. a) shows the impact of fusing parameter $\beta$ in Equ. \ref {eqn:fus}. b) shows the performance of our model with varying the number of images per person in a given batch. c) and d) illustrate the similarity between the probe and gallery images obtained from the baseline and DCDS method, respectively. It can be observed that the baseline method has given larger similarity values for false positive samples (red asterisks above the blue dashed-line) and smaller similarity values for false negative samples (green circles below the blue dashed- line). On the other hand, the proposed DCDS has efficiently assigned the appropriate similarity scores to the true positive and negative samples. 
	}
	\label{fig:paramAnal}
	\vspace{-0.1cm}
\end{figure}

\subsection{Results on Market1501 Datasets}
As can be seen from Table \ref{table:market}, on Market dataset, our proposed method improves state-of-the-art method \cite{KalayehBGKS18}
by $2.5 \%, 1.21 \%,$ and $0.6 \%$ in mAP, rank-1 and rank-5 scores, respectively. Moreover, comparing to state-of-the-art graph-based DNN method, SGGNN \cite{ShenLYCW18DSGNN}, the improvement margins are $3 \%, 2.5 \%, $ and $2 \%$ in mAP, rank-1, and rank-5 score, respectively. Thus, our framework has significantly demonstrated its benefits over state-of-the-art graph-based DNN models. To further improve the result we have adapted a re-ranking scheme \cite{ZhongZCL17}, and we compare our method with state-of-the art methods which use a re-ranking method as a post-processing. As it can be seen from Table \ref*{table:market}, our method has gain mAP of $2.2\%$ over HSP \cite{KalayehBGKS18}, and 10.5 $\%$ over SGGNN\cite{ShenLYCW18DSGNN}, 10.8 $\%$ over DGSRW. 

\subsection{Results on CUHK03 Datasets}
Table \ref{table:compCUHK} shows the performance of our method on CUHK03 dataset. Since most of the Graph-based DNN models report their result on the standard protocol \cite{LiZXW14}, we have experimented on the standard evaluation protocol, to make fair comparison. As can be observed from Table \ref{table:compCUHK}, our method gain a marginal improvement in the mAP. Using a reranking method \cite{ZhongZCL17}, we have reported a competitive result in all evaluation metrics.
\subsection{Results on DukeMTMC-reID Dataset}

Likewise, in DukeMTMC-reID dataset, the improvements of our proposed method is noticeable.
Our method has surpassed state-of-the-art method \cite{KalayehBGKS18} by $1.7\% / 1.6 \% $ in mAP/rank-1 scores. Moreover, comparing to state-of-the-art graph-based DNN, our method outperforms DGSRW \cite{ShenLXYCW18DGSRW},  SGGNN \cite{ShenLYCW18DSGNN} and GCSL \cite{Chen0LSW18} by $9.1 \%, 7.3 \%,$ and $6\%$ in mAP, respectively. 

\subsection{Ablation Study}
To investigate the impact of each component in our architecture, we have performed an ablation study. Thus, we have reported the contributions of each module in Table \ref{table:ABLATION}. To make a fair comparison with the baseline and graph-based DNN models, the ablations study is conducted in a single-dataset (SD) setup.

\textbf{Improvements over the Baseline.} As our main contribution is the DCDS, we examine its impact over the baseline method. The baseline method refers to the lower branch of our architecture that incorporates the verification network, which has also been utilized in \cite{ShenXLYW18DKPM}, \cite{ShenLXYCW18DGSRW}, \cite{ShenLYCW18DSGNN}. On Market1501 dataset, DCDS provides improvements of $9.2 \%, 6.8 \%$ and $3.6 \%$ in mAP, rank-1, and rank-5 scores, respectively, over the baseline method; whereas in DukeMTMC-reID dataset the proposed DCDS improves the baseline method by $8.0 \%, 5.5 \%$ and $1.7 \%$ in mAP, rank-1, and rank-5 scores, respectively.\\
\textbf{Comparison with graph-based deep models.} We  compare our method with recent graph-based-deep models, which adapt similar baseline method as ours, such as \cite{ShenLXYCW18DGSRW},\cite{ShenLYCW18DSGNN}. As a result, on DukeMTMC-reID dataset our method surpass \cite{ShenLXYCW18DGSRW} by $9.1 \% /6.8 \%,$ and \cite{ShenLYCW18DSGNN} by 17.9 $\%$ / 7.4 $\%$ in mAP / rank-1 scores. In light of this, We can   conclude that incorporating a constrained-clustering  mechanism in end-to-end learning has a significant benefit on finding a robust similarity ranking. In addition, experimental findings demonstrate the superiority of DCDS over existing graph-based DNN models.\\
\textbf{Parameter analysis.} 
Experimental results by varying several parameters are shown in Figure \ref{fig:paramAnal}. Figure \ref{fig:paramAnal}(a) shows the effect of fusing parameter, $\beta,$ Equ. (\ref {eqn:fus}) on the mAP. Thereby, we can observe that the mAP tends to increase with a larger $\beta$ value. This shows that the result gets better when we deviate much from the CDS branch. Figure \ref{fig:paramAnal}(b) shows the impact of the number of images per person-ID ($\Omega$) in a given batch. We have experimented setting $\Omega$ to 4, 8, and 16, as can be seen, we obtain  a marginal improvement when we set $\Omega$ to 16. However, considering the direct relationship between the running time and $\Omega$, the improvement is negligible. c) and d) show probe-gallery similarity obtained from baseline and DCDS method, using three different probe-images, with a batch size of 64, and setting $\Omega$ to 4, 8 and 16. 

\section{Conclusion}

In this work, we  presented a novel insight to enhance the learning capability of a DNN through the exploitation of a constrained clustering mechanism. To validate our method, we have conducted extensive experiments on several benchmark datasets. Thereby, the proposed method not only improves state-of-the-art person re-id methods but also demonstrates the benefit of incorporating a constrained-clustering mechanism in the end-to-end learning process. Furthermore, the presented work could naturally be extended to other applications which leverage a similarity-based learning. 
As a future work, we would like to investigate dominant sets clustering as a loss function. 

\section*{Acknowledgment}
This research is partly supported by the Office of the Director of National
Intelligence (ODNI), Intelligence Advanced Research Projects Activity (IARPA), via IARPA
$R\&D$ Contract No. D17PC00345. The views and conclusions contained herein are those
of the authors and should not be interpreted as necessarily representing the official
policies or endorsements, either expressed or implied, of the ODNI, IARPA, or the U.S.
Government. The U.S. Government is authorized to reproduce and distribute reprints
for Governmental purposes notwithstanding any copyright annotation thereon.”

{\small
	\bibliographystyle{ieee}
	\bibliography{egbib}

\begin{thebibliography}{10}\itemsep=-1pt

\bibitem{AhmedJM15}
E.~Ahmed, M.~J. Jones, and T.~K. Marks.
\newblock An improved deep learning architecture for person re-identification.
\newblock In {\em {IEEE} Conference on Computer Vision and Pattern Recognition,
  {CVPR} 2015, Boston, MA, USA, June 7-12, 2015}, pages 3908--3916, 2015.

\bibitem{AlemuPelillo}
L.~T. Alemu and M.~Pelillo.
\newblock Multi-feature fusion for image retrieval using constrained dominant
  sets.
\newblock {\em CoRR}, abs/1808.05075, 2018.

\bibitem{BaiBT17}
S.~Bai, X.~Bai, and Q.~Tian.
\newblock Scalable person re-identification on supervised smoothed manifold.
\newblock In {\em 2017 {IEEE} Conference on Computer Vision and Pattern
  Recognition, {CVPR} 2017, Honolulu, HI, USA, July 21-26, 2017}, pages
  3356--3365, 2017.

\bibitem{BaiZWBLT17}
S.~Bai, Z.~Zhou, J.~Wang, X.~Bai, L.~J. Latecki, and Q.~Tian.
\newblock Ensemble diffusion for retrieval.
\newblock In {\em {IEEE} International Conference on Computer Vision, {ICCV}
  2017, Venice, Italy, October 22-29, 2017}, pages 774--783, 2017.

\bibitem{BertasiusTYS17}
G.~Bertasius, L.~Torresani, S.~X. Yu, and J.~Shi.
\newblock Convolutional random walk networks for semantic image segmentation.
\newblock In {\em 2017 {IEEE} Conference on Computer Vision and Pattern
  Recognition, {CVPR} 2017, Honolulu, HI, USA, July 21-26, 2017}, pages
  6137--6145, 2017.

\bibitem{ChangHX18}
X.~Chang, T.~M. Hospedales, and T.~Xiang.
\newblock Multi-level factorisation net for person re-identification.
\newblock In {\em 2018 {IEEE} Conference on Computer Vision and Pattern
  Recognition, {CVPR} 2018, Salt Lake City, UT, USA, June 18-22, 2018}, pages
  2109--2118, 2018.

\bibitem{Chen0LSW18}
D.~Chen, D.~Xu, H.~Li, N.~Sebe, and X.~Wang.
\newblock Group consistent similarity learning via deep {CRF} for person
  re-identification.
\newblock In {\em 2018 {IEEE} Conference on Computer Vision and Pattern
  Recognition, {CVPR} 2018, Salt Lake City, UT, USA, June 18-22, 2018}, pages
  8649--8658, 2018.

\bibitem{ChenCZH17}
W.~Chen, X.~Chen, J.~Zhang, and K.~Huang.
\newblock Beyond triplet loss: {A} deep quadruplet network for person
  re-identification.
\newblock In {\em {CVPR}}, pages 1320--1329. {IEEE} Computer Society, 2017.

\bibitem{ChengGZWZ16}
D.~Cheng, Y.~Gong, S.~Zhou, J.~Wang, and N.~Zheng.
\newblock Person re-identification by multi-channel parts-based {CNN} with
  improved triplet loss function.
\newblock In {\em 2016 {IEEE} Conference on Computer Vision and Pattern
  Recognition, {CVPR} 2016, Las Vegas, NV, USA, June 27-30, 2016}, pages
  1335--1344, 2016.

\bibitem{dgray}
S.~B. D.~Gray and H.~Tao.
\newblock Evaluating appearance models for recognition, reacquisition.
\newblock In {\em IEEE International workshop on performance evaluation of
  track- ing and surveillance,}, pages 31--44, 2007.

\bibitem{DingLWC15}
S.~Ding, L.~Lin, G.~Wang, and H.~Chao.
\newblock Deep feature learning with relative distance comparison for person
  re-identification.
\newblock {\em Pattern Recognition}, 48(10):2993--3003, 2015.

\bibitem{DonoserB13}
M.~Donoser and H.~Bischof.
\newblock Diffusion processes for retrieval revisited.
\newblock In {\em {CVPR}}, pages 1320--1327. {IEEE} Computer Society, 2013.

\bibitem{FanZYY18}
H.~Fan, L.~Zheng, C.~Yan, and Y.~Yang.
\newblock Unsupervised person re-identification: Clustering and fine-tuning.
\newblock {\em {TOMCCAP}}, 14(4):83:1--83:18, 2018.

\bibitem{FelzenszwalbGMR10}
P.~F. Felzenszwalb, R.~B. Girshick, D.~A. McAllester, and D.~Ramanan.
\newblock Object detection with discriminatively trained part-based models.
\newblock {\em {IEEE} Trans. Pattern Anal. Mach. Intell.}, 32(9):1627--1645,
  2010.

\bibitem{HeZRS16}
K.~He, X.~Zhang, S.~Ren, and J.~Sun.
\newblock Deep residual learning for image recognition.
\newblock In {\em {CVPR}}, pages 770--778. {IEEE} Computer Society, 2016.

\bibitem{KalayehBGKS18}
M.~M. Kalayeh, E.~Basaran, M.~G{\"{o}}kmen, M.~E. Kamasak, and M.~Shah.
\newblock Human semantic parsing for person re-identification.
\newblock In {\em 2018 {IEEE} Conference on Computer Vision and Pattern
  Recognition, {CVPR} 2018, Salt Lake City, UT, USA, June 18-22, 2018}, pages
  1062--1071, 2018.

\bibitem{KipfW16}
T.~N. Kipf and M.~Welling.
\newblock Semi-supervised classification with graph convolutional networks.
\newblock {\em CoRR}, abs/1609.02907, 2016.

\bibitem{LiZW12}
W.~Li, R.~Zhao, and X.~Wang.
\newblock Human reidentification with transferred metric learning.
\newblock In {\em Computer Vision - {ACCV} 2012 - 11th Asian Conference on
  Computer Vision, Daejeon, Korea, November 5-9, 2012, Revised Selected Papers,
  Part {I}}, pages 31--44, 2012.

\bibitem{LiZXW14VFn}
W.~Li, R.~Zhao, T.~Xiao, and X.~Wang.
\newblock Deepreid: Deep filter pairing neural network for person
  re-identification.
\newblock In {\em 2014 {IEEE} Conference on Computer Vision and Pattern
  Recognition, {CVPR} 2014, Columbus, OH, USA, June 23-28, 2014}, pages
  152--159, 2014.

\bibitem{LiZXW14}
W.~Li, R.~Zhao, T.~Xiao, and X.~Wang.
\newblock Deepreid: Deep filter pairing neural network for person
  re-identification.
\newblock In {\em {CVPR}}, pages 152--159. {IEEE} Computer Society, 2014.

\bibitem{LiZG18}
W.~Li, X.~Zhu, and S.~Gong.
\newblock Harmonious attention network for person re-identification.
\newblock In {\em {CVPR}}, pages 2285--2294. {IEEE} Computer Society, 2018.

\bibitem{LoyLG13}
C.~C. Loy, C.~Liu, and S.~Gong.
\newblock Person re-identification by manifold ranking.
\newblock In {\em {IEEE} International Conference on Image Processing, {ICIP}
  2013, Melbourne, Australia, September 15-18, 2013}, pages 3567--3571, 2013.

\bibitem{LoyXG09}
C.~C. Loy, T.~Xiang, and S.~Gong.
\newblock Multi-camera activity correlation analysis.
\newblock In {\em {CVPR}}, pages 1988--1995. {IEEE} Computer Society, 2009.

\bibitem{PavPel07}
M.~Pavan and M.~Pelillo.
\newblock Dominant sets and pairwise clustering.
\newblock {\em {IEEE} Trans. Pattern Anal. Mach. Intell.}, 29(1):167--172,
  2007.

\bibitem{ShenLXYCW18DGSRW}
Y.~Shen, H.~Li, T.~Xiao, S.~Yi, D.~Chen, and X.~Wang.
\newblock Deep group-shuffling random walk for person re-identification.
\newblock In {\em 2018 {IEEE} Conference on Computer Vision and Pattern
  Recognition, {CVPR} 2018, Salt Lake City, UT, USA, June 18-22, 2018}, pages
  2265--2274, 2018.

\bibitem{ShenLYCW18DSGNN}
Y.~Shen, H.~Li, S.~Yi, D.~Chen, and X.~Wang.
\newblock Person re-identification with deep similarity-guided graph neural
  network.
\newblock In {\em Computer Vision - {ECCV} 2018 - 15th European Conference,
  Munich, Germany, September 8-14, 2018, Proceedings, Part {XV}}, pages
  508--526, 2018.

\bibitem{ShenXLYW18DKPM}
Y.~Shen, T.~Xiao, H.~Li, S.~Yi, and X.~Wang.
\newblock End-to-end deep kronecker-product matching for person
  re-identification.
\newblock In {\em 2018 {IEEE} Conference on Computer Vision and Pattern
  Recognition, {CVPR} 2018, Salt Lake City, UT, USA, June 18-22, 2018}, pages
  6886--6895, 2018.

\bibitem{SuhWTML18}
Y.~Suh, J.~Wang, S.~Tang, T.~Mei, and K.~M. Lee.
\newblock Part-aligned bilinear representations for person re-identification.
\newblock In {\em Computer Vision - {ECCV} 2018 - 15th European Conference,
  Munich, Germany, September 8-14, 2018, Proceedings, Part {XIV}}, pages
  418--437, 2018.

\bibitem{TesfayeZPP16}
Y.~T. Tesfaye, E.~Zemene, M.~Pelillo, and A.~Prati.
\newblock Multi-object tracking using dominant sets.
\newblock {\em {IET} Computer Vision}, 10(4):289--297, 2016.

\bibitem{TesfayeZPPS17}
Y.~T. Tesfaye, E.~Zemene, A.~Prati, M.~Pelillo, and M.~Shah.
\newblock Multi-target tracking in multiple non-overlapping cameras using
  constrained dominant sets.
\newblock {\em CoRR}, abs/1706.06196, 2017.

\bibitem{VariorHW16}
R.~R. Varior, M.~Haloi, and G.~Wang.
\newblock Gated siamese convolutional neural network architecture for human
  re-identification.
\newblock In {\em Computer Vision - {ECCV} 2016 - 14th European Conference,
  Amsterdam, The Netherlands, October 11-14, 2016, Proceedings, Part {VIII}},
  pages 791--808, 2016.

\bibitem{WangPS17}
C.~Wang, M.~Pelillo, and K.~Siddiqi.
\newblock Dominant set clustering and pooling for multi-view 3d object
  recognition.
\newblock In {\em {BMVC}}. {BMVA} Press, 2017.

\bibitem{WangCW018}
Y.~Wang, Z.~Chen, F.~Wu, and G.~Wang.
\newblock Person re-identification with cascaded pairwise convolutions.
\newblock In {\em 2018 {IEEE} Conference on Computer Vision and Pattern
  Recognition, {CVPR} 2018, Salt Lake City, UT, USA, June 18-22, 2018}, pages
  1470--1478, 2018.

\bibitem{WangWYZCLHHW18}
Y.~Wang, L.~Wang, Y.~You, X.~Zou, V.~Chen, S.~Li, G.~Huang, B.~Hariharan, and
  K.~Q. Weinberger.
\newblock Resource aware person re-identification across multiple resolutions.
\newblock In {\em 2018 {IEEE} Conference on Computer Vision and Pattern
  Recognition, {CVPR} 2018, Salt Lake City, UT, USA, June 18-22, 2018}, pages
  8042--8051, 2018.

\bibitem{WeiZ0018}
L.~Wei, S.~Zhang, W.~Gao, and Q.~Tian.
\newblock Person transfer {GAN} to bridge domain gap for person
  re-identification.
\newblock In {\em {CVPR}}, pages 79--88. {IEEE} Computer Society, 2018.

\bibitem{Wei95}
J.~W. Weibull.
\newblock {\em Evolutionary Game Theory}.
\newblock MIT press, 1995.

\bibitem{YanXL18}
S.~Yan, Y.~Xiong, and D.~Lin.
\newblock Spatial temporal graph convolutional networks for skeleton-based
  action recognition.
\newblock In {\em Proceedings of the Thirty-Second {AAAI} Conference on
  Artificial Intelligence, New Orleans, Louisiana, USA, February 2-7, 2018},
  pages 7444--7452, 2018.

\bibitem{ZemeneAP16}
E.~Zemene, L.~T. Alemu, and M.~Pelillo.
\newblock Constrained dominant sets for retrieval.
\newblock In {\em 23rd International Conference on Pattern Recognition, {ICPR}
  2016, Canc{\'{u}}n, Mexico, December 4-8, 2016}, pages 2568--2573, 2016.

\bibitem{ZemeneAP17}
E.~Zemene, L.~T. Alemu, and M.~Pelillo.
\newblock Dominant sets for "constrained" image segmentation.
\newblock {\em CoRR}, abs/1707.05309, 2017.

\bibitem{ZemenePECCV16}
E.~Zemene and M.~Pelillo.
\newblock Interactive image segmentation using constrained dominant sets.
\newblock In {\em Computer Vision - {ECCV} 2016 - 14th European Conference,
  Amsterdam, The Netherlands, October 11-14, 2016, Proceedings, Part {VIII}},
  pages 278--294, 2016.

\bibitem{ZemeneTIPPS19}
E.~Zemene, Y.~T. Tesfaye, H.~Idrees, A.~Prati, M.~Pelillo, and M.~Shah.
\newblock Large-scale image geo-localization using dominant sets.
\newblock {\em {IEEE} Trans. Pattern Anal. Mach. Intell.}, 41(1):148--161,
  2019.

\bibitem{ZhaoLZW17}
L.~Zhao, X.~Li, Y.~Zhuang, and J.~Wang.
\newblock Deeply-learned part-aligned representations for person
  re-identification.
\newblock In {\em {IEEE} International Conference on Computer Vision, {ICCV}
  2017, Venice, Italy, October 22-29, 2017}, pages 3239--3248, 2017.

\bibitem{ZhengSTWWT15}
L.~Zheng, L.~Shen, L.~Tian, S.~Wang, J.~Wang, and Q.~Tian.
\newblock Scalable person re-identification: {A} benchmark.
\newblock In {\em {ICCV}}, pages 1116--1124. {IEEE} Computer Society, 2015.

\bibitem{ZhengGX09}
W.~Zheng, S.~Gong, and T.~Xiang.
\newblock Associating groups of people.
\newblock In {\em {BMVC}}, pages 1--11. British Machine Vision Association,
  2009.

\bibitem{ZhengZY17}
Z.~Zheng, L.~Zheng, and Y.~Yang.
\newblock Unlabeled samples generated by {GAN} improve the person
  re-identification baseline in vitro.
\newblock In {\em {ICCV}}, pages 3774--3782. {IEEE} Computer Society, 2017.

\bibitem{ZhongZCL17}
Z.~Zhong, L.~Zheng, D.~Cao, and S.~Li.
\newblock Re-ranking person re-identification with k-reciprocal encoding.
\newblock In {\em {CVPR}}, pages 3652--3661. {IEEE} Computer Society, 2017.

\bibitem{Zhong}
Z.~Zhong, L.~Zheng, G.~Kang, S.~Li, and Y.~Yang.
\newblock Random erasing data augmentation.
\newblock {\em CoRR}, abs/1708.04896, 2017.

\end{thebibliography}
}

\begin{appendices}
		
\begin{figure}
		\begin{center}
			
			\includegraphics[width=1\linewidth ,trim=0cm 0cm 0cm 0cm,clip]{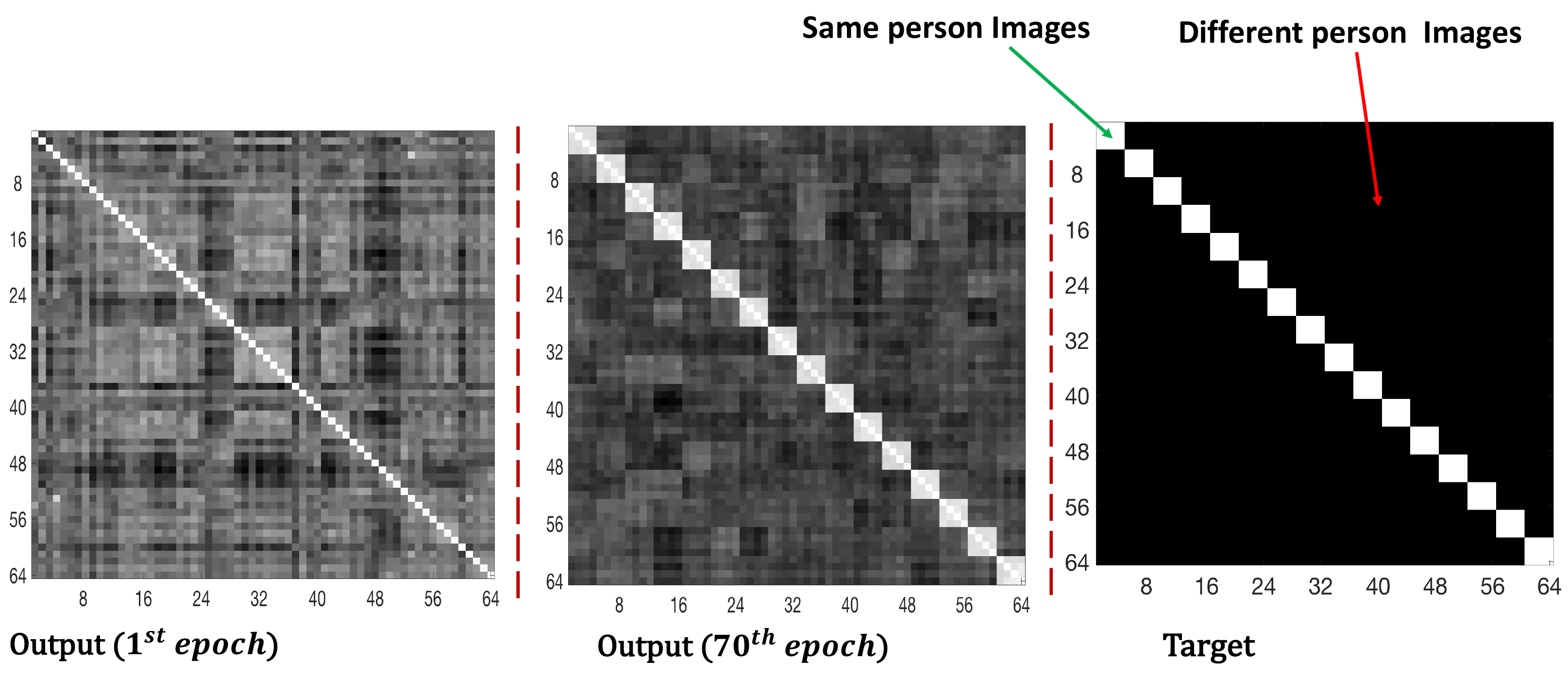}
		\end{center}
		
		\caption{On the right hand side, the target matrix is shown.
			There are total 16 persons in the mini-batch and 4 images per ID ($\Omega$ = 4),  batch size = 64. In the target matrix, the white-blocks represent the similarity between the same person-images in the mini-batch, whereas the black-blocks of the matrix define the dissimilarities between different person images. In the similarity matrix shown left ( after one epoch) and middle (after $70^{th}$ epochs) each row of the output matrix denotes the fused similarity obtained from the CDS-Net and V-Net, per Equ. (6) in the main manuscript. Thus, we optimize our model until we obtain an output with a similar distribution of the target matrix. As can be seen, our model has effectively learned and gives a similarity matrix (shown in the middle) which is closer to the target matrix. }
		\label{fig:metaphore}
		
	\end{figure}

\begin{figure*}
		\begin{center}
			
			\includegraphics[width=1\linewidth ,trim=0cm 0cm 0cm 0cm,clip]{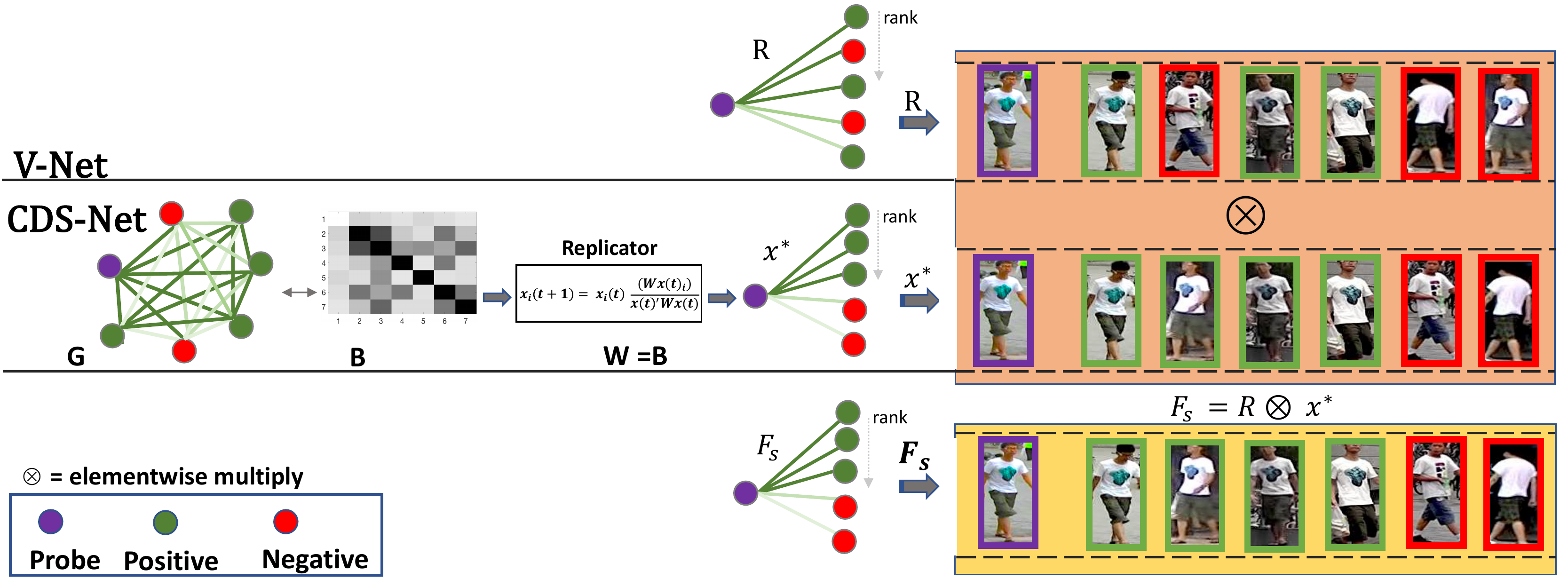}
		\end{center}
		
		\caption{Exemplar results obtained as a result of the similarity fusion between the V-Net and CDS-Net. The Upper-row shows the probe and gallery similarity (R) obtained from the V-Net, where the green circles show persons similar to the probe (shown by purple-circle), while the red circles denote persons different from the probe image. Middle-row shows the workflow in CDS-Net. First, graph G is formed using the similarity obtained from the dot products. We then construct the modified affinity matrix $B$, followed by application of replicator dynamics on $B$ to obtain the probe gallery similarity ($X^*$). Finally, We elementwise multiply $X^*$ and $R$ to find the final probe-gallery similarity ($F_s$), shown in the third row. The intensity of the edges in, $G,$ $R,$ $x^*,$ and $F_s$ define the similarity value, where the bold ones denote larger similarity values, whereas the pale-edges depict smaller similarity values.}
		\label{fig:exemplarResult}
		
	\end{figure*}

	In the supplementary material, we provide additional experiments on cross-dataset person-re-identification (re-id) using the proposed deep constrained dominant sets (DCDS) on Market1501 dataset. In section one, we summarize the  datasets we used in our experiments. In section two, we present the experiments we have performed on cross-dataset person re-id. And, in section three, we provide hyper  parameter analysis on DukeMTMC-reID and CUHK03 datasets. Figure \ref{fig:metaphore} illustrates an example of our method training-output (left) and learning objective, target matrix, (right). Figure \ref{fig:exemplarResult} demonstrates the similarity fusing process, between the V-Net and CDS-Net, alongside sample qualitative results.
	
	\section{Datasets}
	
	
	In multiple dataset (MD) setup, we first train our model on eight datasets: CUHK03 \cite{LiZXW14}, CUHK01 \cite{LiZW12}, Market1501 \cite{ZhengSTWWT15}, DukeMTMC-reID \cite{ZhengZY17}, Viper \cite{dgray}, MSMT17 \cite{WeiZ0018}, GRID \cite{LoyXG09}, and ILIDS \cite{ZhengGX09}. Next, we fine-tune and evaluate on each of   CUHK03 \cite{LiZXW14}, Market1501 \cite{ZhengSTWWT15}, and DukeMTMC-reID \cite{ZhengZY17} datasets.
	
	\section{Experiments on Cross-datasets Evaluation}

	Due to the lack of abundant labeled data, cross-dataset person re-id has attracted great interest. Recently, Fan \et\cite{FanZYY18} have developed a progressive clustering-based method to attack cross-dataset person re-id problem. To further validate our proposed DCDS, we apply our method on cross-dataset person re-id problem and compare it with progressive unsupervised learning (PUL)  \cite{FanZYY18}. To this end, we train our model on DukeMTMC-reID and CUHK03 datasets and test it on Market1501 dataset.  We then compare it with PUL \cite{FanZYY18}, which has also been trained on CUHK03 and DukeMTMC-reID datasets. As can be observed from Table \ref{table:compCUHK}, even though our proposed method is not intended for cross-dataset re-id, it has gained a substantial improvements over PUL \cite{FanZYY18}, that was  mainly designed to attack person re-id problem in a cross-dataset setup.

	\section{Parameter Analysis}
\begin{figure*}
		\begin{center}
			
			\includegraphics[width=1\linewidth ,trim=0cm 0cm 0cm 0cm,clip]{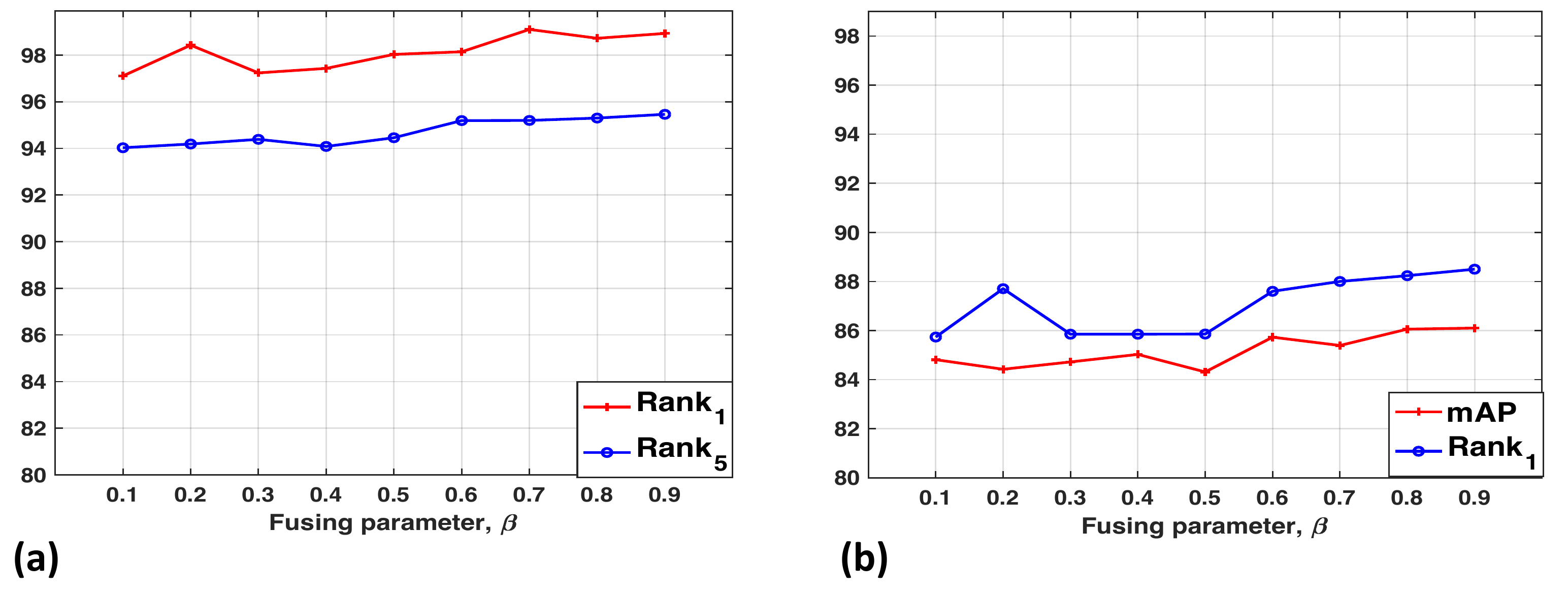}
		\end{center}
		
		\caption{Performance of our model with respect to fusing parameter $\beta$, on (a) CUHK03,  and   (b) DukeMTMC-reID, datasets.}
		\label{fig:fusingpar}
		
	\end{figure*}

\begin{figure*}
		\vspace{-0.3cm}
		
		\begin{center}
			
			\includegraphics[width=1\linewidth ,trim=0cm 0cm 0cm 0cm,clip]{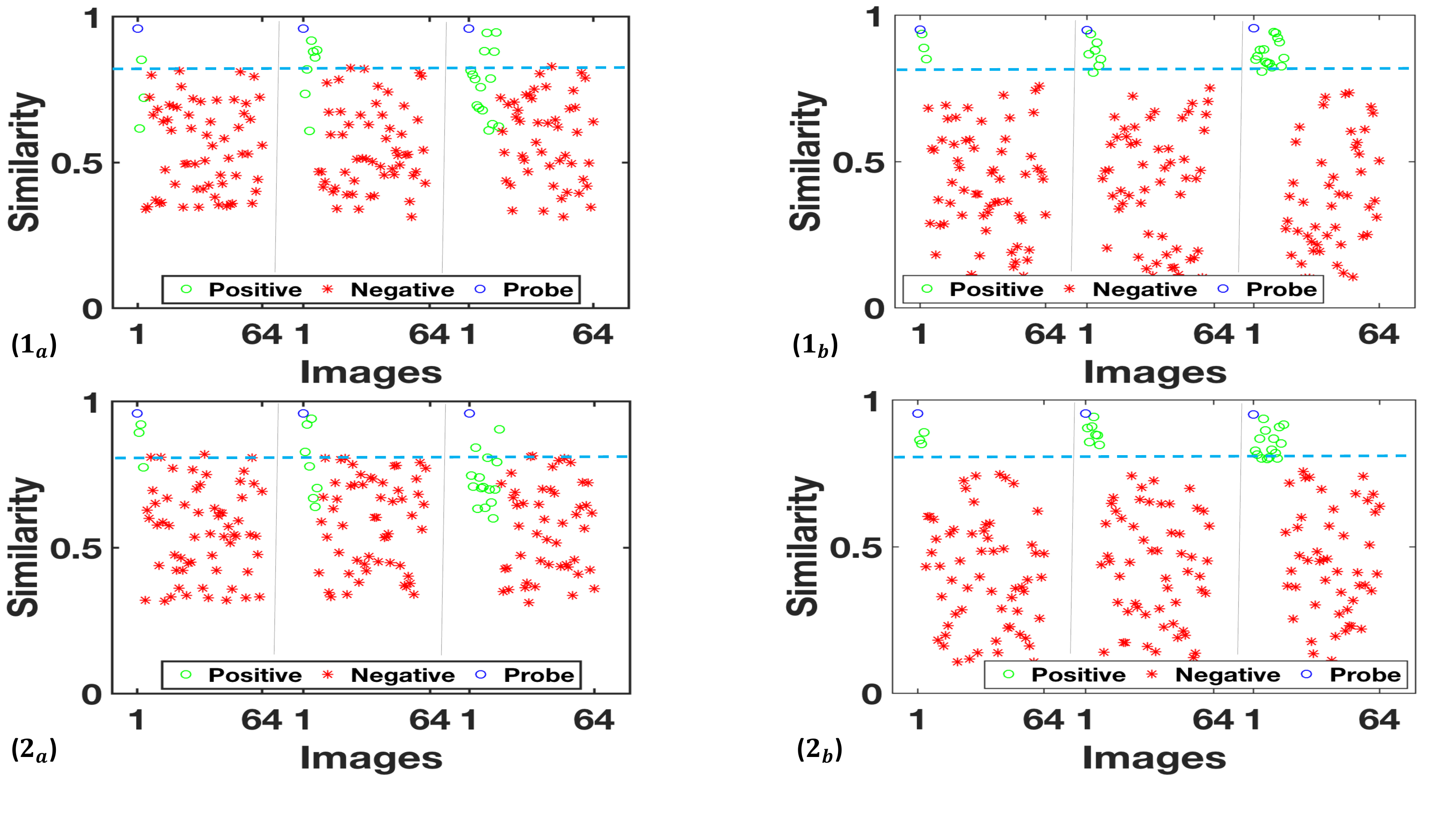}
		\end{center}
		
		\caption{Shows experimental analysis performed on CUHK03 $(1_{a, b}),$ and DukeMTMC-reID $(2_{a, b})$ datasets. $1_a, 2_a$ and $1_b, 2_b$ illustrate the similarity between the probe-gallery images obtained from the baseline and DCDS method, respectively. It can be observed that the baseline method has assigned larger similarity values for false positive samples (red asterisks above the blue dashed-line) and smaller similarity values for false negative samples (green circles below the blue dashed-line). On the other hand, the proposed DCDS has efficiently assigned the appropriate similarity scores to the true positive and negative samples. Note that, for better visibility, we have randomly assigned a large (close to 1) self-similarity value to the probe (blue-circle).}
		\label{fig:SimilDist}
		
	\end{figure*}

	Similar to  the parameter analysis reported in the main manuscript, we report hyper parameter analysis on DukeMTMC-reID and CUHK03 dataset. The performance of our method with respect to the fusing parameters on DukeMTMC-reID and CUHK03 are shown in Figure \ref{fig:fusingpar} (a) and Figure \ref{fig:fusingpar} (b), respectively. Thereby, as can be observed, the results  show similar phenomena as in Market1501, where the mAP increases with a larger $\beta$ value. Figure \ref{fig:SimilDist} shows the similarity distribution given by the baseline and the proposed DCDS using three different probe-images, with a batch size of 64, and setting $\Omega$ to 4, 8 and 16.
	
	
	\begin{table}
		
		\begin{center}
			
			\begin{tabular}{l|l|c} 
				\hline
				
				&\multicolumn{2}{l}{\small {\bf Train} on Duke, CUHK03 $\rightarrow$ {\bf Test} on Market1501}\\
				\hline
				Methods & mAP & rank-1\\
				
				\hline\hline
				PUL \cite{FanZYY18}  & 20.5 &45.5   \\
				Ours   & \textbf{24.5} & \textbf{51.3}   \\
				\hline
			\end{tabular}
		\end{center}
		
		\caption{A comparison of the proposed method with PUL \cite{FanZYY18} on Market1501 dataset.}
		\label{table:compCUHK}
		
	\end{table}
	
\end{appendices}


\end{document}